\pdfoutput=1

\documentclass[journal]{IEEEtran}
\usepackage{cite}

\ifCLASSINFOpdf
  \usepackage[pdftex]{graphicx}
  \graphicspath{{../pdf/}{../jpeg/}}
   \DeclareGraphicsExtensions{.pdf,.jpeg,.png}
\else
   \usepackage[dvips]{graphicx}
   \graphicspath{{../eps/}}
   \DeclareGraphicsExtensions{.eps}
\fi
\ifCLASSOPTIONcompsoc
\usepackage[caption=false,font=normalsize,labelfon
t=sf,textfont=sf]{subfig}
\else
\usepackage[caption=false,font=footnotesize]{subfi
	g}
\fi

\usepackage{amsmath}
\usepackage{amsfonts} 
\usepackage{algorithm}
\usepackage{algorithmic}

\usepackage{booktabs} 

\begin{document}

%
\title{Boosting Independent Component Analysis}
%
%
%

\author{YunPeng ~Li\quad ZhaoHui ~Ye
	
	\thanks{YunPeng Li (corresponding author) is with the D-ITET, ETH Zurich, Zurich, Switzerland e-mail: yunpli@isi.ee.ethz.ch}
	\thanks{ZhaoHui Ye is with the Department of Automation, Tsinghua University, Beijing, China e-mail: yezhaohui@mail.tsinghua.edu.cn}
}

%
%

\markboth{Journal of \LaTeX\ Class Files,~Vol.~14, No.~8, August~2015}%
{Shell \MakeLowercase{\textit{et al.}}: Bare Demo of IEEEtran.cls for IEEE Journals}
%



\maketitle

\begin{abstract}
Independent component analysis is intended to recover the mutually independent components from their linear mixtures. This technique has been widely used in many fields, such as data analysis, signal processing, and machine learning. To alleviate the dependency on prior knowledge concerning unknown sources, many nonparametric methods have been proposed. In this paper, we present a novel boosting-based algorithm for independent component analysis. Our algorithm consists of maximizing likelihood estimation via boosting and seeking unmixing matrix by the fixed-point method. A variety of experiments validate its performance compared with many of the presently known algorithms.
\end{abstract}


%
\IEEEpeerreviewmaketitle

\section{Introduction}
%
%
%
%
\IEEEPARstart{I}{ndependent} Component Analysis (ICA)  has received much attention in recent years due to its effective methodology for various problems, such as feature extraction, blind source separation, and exploratory data analysis. In the ICA model, a $m$ dimensional random vector $\mathbf{x}=(x_{1},\cdots,x_{m})^{T}$ is observed as a linear mixture of the $m$ independent sources $\mathbf{s}=(s_{1},\cdots,s_{m})^{T}$,
\begin{equation}
\label{mixing}
\mathbf{x}=\mathbf{A}\mathbf{s}
\end{equation}
where $\mathbf{s}$ contains the mutually independent components, and $\mathbf{A}$ is called the mixing matrix. An unmixing matrix $\mathbf{W}$ is used to determine the estimated sources $\mathbf{y}=(y_{1},\cdots,y_{m})^{T}$
from the observed mixtures $\mathbf{x}$, where the source's estimation $\mathbf{y}$ is the scaling and permutation of the original source $\mathbf{s}$. The recovering process can be modeled as
\begin{equation}
\label{unmixing}
\mathbf{y}=\mathbf{W}\mathbf{x}
\end{equation}
Without loss of generality, several assumptions are clear in ICA : (i) $\mathrm{E}(\mathbf{s})=\mathbf{0}$ and $\mathrm{Cov}(\mathbf{s})=\mathbf{I}$; (ii) each $s_{i}$  is independent distributed and at most one of the sources is Gaussian \cite{Comon1994}; (iii) both $\mathbf{A}$ and $\mathbf{W}$ are invertible matrices. Since centering and whitening are often performed as preprocessing steps in the context of linear ICA, $\mathbf{W}$ is restricted to be an orthonormal matrix $\mathbf{W}\mathbf{W}^{T}=\mathbf{I}$.
 
ICA was firstly introduced to the neural network domain in the 1980s\cite{HJ1986}. It was not until 1994 \cite{Comon1994} that the theory of ICA was established. The most popular method is to optimize some contrast functions to achieve source separation. These contrast functions were usually chosen to represent the measure of independence or non-Gaussianity, for example, the mutual information \cite{BellSej1995,Lee1999}, the maximum entropy (or negentropy) \cite{fastica1999,RADICAL,ypl2022}, and the nonlinear decorrelation \cite{JUTTEN19911,Bach2003}. In addition, higher-order moments methods \cite{JCar1989,JCar1993} were also designed to estimate the unknown sources. It has been pointed out that these contrast functions are related to the sources' density distributions in the parametric maximum likelihood estimation \cite{Pear1996,mky1996,DTP1997,JC1997}. These parametric methods' performance are dependent on the choices of contrast functions or prior assumptions on the unknown sources' distributions. Although it\cite{fastica1999} has been shown that small misspecifications (for parametric methods) in the source densities do not affect local consistency of the estimation concerning the unmixing matrix $\mathbf{W}$, Amari\cite{amari2002independent} demonstrated that ICA methods have to estimate the sources' densities (nonparametric methods) to achieve the information bound concerning $\mathbf{W}$. 

Nonparametric ICA's in-depth analysis and asymptotic efficiency were firstly available in \cite{chen2006}. There are currently two kinds of nonparametric ICA methods: the restriction methods and the regularization methods. In restriction methods, each source $s_{i}$ belongs to certain density family or owns special structure, such as Gaussian mixtures models \cite{EM1997,welling2001}, kernel density distributions \cite{Nikos2001,Eloyan2013}, and log-concave family \cite{Sam2012}. The regularization methods determine the unknown sources via  maximizing a penalized likelihood \cite{HaTi2002}. For other recent ICA approaches, see also\cite{Ilmonen2011,palmer2012amica,Matteson2017,Pierre2018,spurek2018,Anastasia2019}.

Recently, we have successfully applied boosting to the nonparametric maximum likelihood estimation (boosting NPMLE) \cite{ypl2021}. In this paper, we further propose a novel ICA method called $BoostingICA$ based on our previous work. The proposed approach  adaptively includes only those basis functions that contribute significantly to the sources' estimation. Real data experiments validate its competitive performance with other popular or recent ICA methods.

\section{Nonparametric maximum likelihood estimation in ICA}
\label{sec2}
In this section, nonparametric maximum likelihood estimation is applied to solve the ICA problem. Maximizing the likelihood can be viewed as a joint maximization over the unmixing matrix $\mathbf{W}$ and the estimated sources' densities, fixing one argument and maximizing over the other. 

Let $p_{i}(y_{i})$ be the estimated density distribution for single component $y_{i}$. Owing to the independence among $y_{i}$, the estimated sources' density distribution $p_{\mathbf{y}}(\mathbf{y})$ can be written as
\begin{equation}
\label{likelihood}
p_{\mathbf{y}}(\mathbf{y}) =\prod_{i=1}^{m}p_{i}(y_{i})
\end{equation}
Since there is a linear transform in equation (\ref{unmixing}), the joint density of the observed mixtures $\mathbf{x}$ is 
\begin{equation}
\label{likelihood2}
p_{\mathbf{x}}(\mathbf{x};\mathbf{W},\mathbf{y}) =|\mathbf{det(W)})|p_{\mathbf{y}}(\mathbf{W}\mathbf{x})\\
=\prod_{i=1}^{m}p_{i}(\mathbf{w}^{T}_{i}\mathbf{x})\\
=\prod_{i=1}^{m}p_{i}(y_{i})
\end{equation}
where $\mathbf{w}_{i}^{T}$ is the $i_{th}$ row in the orthonormal matrix $\mathbf{W}$ and $|\mathbf{det(W)})|=1$.

For parametric ICA, We have to determine the density distribution $p_{i}(y_{i})$ based on the prior assumptions beforehand, which might result in the unwanted density mismatching. To keep the density distribution of $y_{i}$ unspecified, nonparametric methods are demanded. In the proposed method, we choose to model the estimated source $y_{i}$ with Gibbs distribution in equation (\ref{gibbs}).

\begin{equation}
\label{gibbs}
p_{i}(y_{i})=\frac{e^{f_{i}(y_{i})}}{\int e^{f_{i}(y_{i})}dy_{i}}
\end{equation}
where $f_{i}(y_{i})$ is assumed to be a smooth function in $\mathbb{R}$, and the denominator of equation (\ref{gibbs}) is the partition function. Given $N$ independent identically distributed samples $\{\mathbf{x}_{j}\}_{1}^{N}$, we can estimate the source's sample $y_{i}^{j}=\mathbf{w}_{i}^{T}\mathbf{x}_{j}$. 

To simplify the integral in equation (\ref{gibbs}), we construct a grid of $L$ values $y_{i}^{*l}$ (ascending order) with $\Delta_{i}$ step , and let the corresponding frequency $q_{i}^{l}$ be
\begin{equation}
\label{eq:freq}
q_{i}^{l} = \sum_{j=1}^{N}\mathbb{I}(y_{i}^{j}\in \left(y_{i}^{*l}-\Delta_{i}/2,y_{i}^{*l}+\Delta_{i}/2\right])/N
\end{equation}
where $\mathbb{I}(.)$ is the indicator function. The support of $p_{i}(y_{i})$ is restricted in $[y_{i}^{*1}-\Delta_{i}/2,y_{i}^{*L}+\Delta_{i}/2]$, and $p_{i}(y_{i})$ is then transformed as
\begin{equation}
\label{gibbs2}
p_{i}(y_{i})=e^{f_{i}(y_{i})}/\sum_{l=1}^{L} \Delta_{i} e^{f_{i}(y_{i}^{*l})}
\end{equation}
It \cite{silverman1982,ypl2021} has been shown that the log-likelihood $\mathcal{L}(\mathbf{w}_{i},f_{i})$ concerning $y_{i}$ can be simplified to the following modified form 
\begin{equation}
\label{loglikelihood1}
\mathcal{L}(\mathbf{w}_{i},f_{i}) =\sum_{l=1}^{L} q_{i}^{l} \ln \,p_{i}(y_{i}^{*l}) =\sum_{l=1}^{L} q_{i}^{l}f_{i}(y_{i}^{*l})-\Delta_{i}e^{f_{i}(y_{i}^{*l})}
\end{equation}
and the maximum log-likelihood is obtained when the partition function $\sum_{l=1}^{L} \Delta_{i}e^{f_{i}(y_{i}^{*l})}=1$.
Thus, the total modified log-likelihood in our method is
\begin{equation}
\label{loglikelihood2}
\mathcal{L}(\{\mathbf{w}_{i},f_{i}\}_{i=1}^{m}) = \sum_{i=1}^{m} \mathcal{L}(\mathbf{w}_{i},f_{i})
\end{equation}

The key of $BoostingICA$ is to optimize equation (\ref{loglikelihood2}) until convergence by joint maximization\cite{HaTi2002,kold2006,Sam2012}, and the two iterative stages are illustrated in Algorithm \ref{alg:ICA}.
\begin{algorithm}[!t]  
	\caption{BoostingICA algorithm}  
	\label{alg:ICA}  
	\begin{algorithmic}[1]
		\STATE{Initialization}  
		\STATE{maximum iterations $maxit$}
		\STATE{grid value $L$}
		\STATE{boosting iterations $M$}
		\STATE{degree of freedom $df$}
		\FOR{$it = 1$ to $maxit$}
		\FOR{$i = 1$ to $m$}
		\STATE{fixing $\mathbf{W}$, each $f_{i}$ is estimated by boosting NPMLE (Algorithm \ref{alg:boosting}).}
		\ENDFOR
		\STATE{given $f_{i}$, $\mathbf{W}$ is restricted to be orthonormal and calculated via the fixed-point method (Algorithm \ref{alg:fixed-point}).}
		\ENDFOR	
		\STATE{Output}
		\STATE{unmixing matrix $\mathbf{W}$}
		\STATE{estimated densities $\{f_{i}\}_{i=1}^{m}$}
	\end{algorithmic}  
\end{algorithm}
Similar joint maximization has been used in past researches concerning both projection pursuit \cite{PPDE1984,EPP1987} and ICA \cite{HaTi2002,Sam2012}.
\subsection{Estimating the source's density via boosting}
\label{sec2_1}
To apply the boosting principle (forward-stagewise regression) \cite{FREUND1997119,Friedman2001a} to nonparametric maximum likelihood estimation\cite{ypl2021}, we regard $f_{i}(y_{i})$ in equation (\ref{loglikelihood1}) as a combination of weak learners $b_{i}(y_{i};\gamma_{i,k})$,
\begin{equation}
\label{boosting1}
f_{i}(y_{i}) =\sum_{k=1}^{M}b_{i}(y_{i};\gamma_{i,k})\\
\end{equation}
where $M$ is the number of boosting iterations and $k$ is the index. Each single weak learner $b_{i}(y_{i};\gamma_{i,k})$ is characterized by a set of parameters $\gamma_{i,k}$  and is trained on the weighted data at the $k_{th}$ iteration. We define the $f_{i}$ at the $k_{th}$ iteration as $f_{i}^{k}$
\begin{equation}
\label{boosting3}
f_{i}^{k}(y_{i})=\sum_{r=1}^{k}b(y_{i};\gamma_{i,r})
=f_{i}^{k-1}(x)+b(y_{i};\gamma_{i,k})
\end{equation}
and the subproblem based on the former $f_{i}^{k-1}$ becomes
\begin{equation}
\label{boosting2}
\max_{\gamma_{i,k}}\quad \mathcal{L}(f_{i}^{k-1}(y_{i})+b(y_{i};\gamma_{i,k}))
\end{equation}
We maximize equation (\ref{boosting2}) via second-order approximation around $f_{i}^{k-1}$, and the corresponding iterative reweighted least squares (IRLS)\cite{IRLS01} (Line 14) and updating strategy for the next iteration (Line 10-11) are shown in Algorithm \ref{alg:boosting}. A penalty term $\lambda J(b_{i}(y_{i};\gamma_{i,k}))$ (Line 14) is added to the original least squares to restrict the model complexity of $b_{i}(y_{i};\gamma_{i,k})$, where $\lambda$ is the Lagrange multiplier and $J(b_{i}(y_{i};\gamma_{i,k}))$ is a nonnegative function. $\omega_{l}^{k}$, $Y_{l}^{k}$ (Line 10-11) are the weight and response of ${y_{i}^{*l}}$ at the $k_{th}$ iteration, and they are simultaneously updated in the next iteration. Once all the weak learners have been trained, $f_{i}(y_{i})$ is the combination of whole $M$ weak learners.

Inspired by the past researches \cite{HaTi2002,ypl2021}, smooth spline is chosen as the weak learner for two reasons: Line 14 in Algorithm \ref{alg:boosting} can be efficiently computed in $\mathcal{O}(L)$ time \cite{HaTi2002}, and then the first and second derivatives of smooth spline (concerning $y_{i}$) are immediately available \cite{bookSpline,HaTi2002}. The penalty term is defined as $J(b_{i}(y_{i};\gamma_{i,k})) = \int\left[b_{i}^{''}(y_{i};\gamma_{i,k})\right]^{2}dy_{i}$ (roughness penalty penalizes the curvature of function), and the degree of freedom \cite{esl} $df$ (corresponding to the prechosen $\lambda$) is defined implicitly by the trace of the linear smoother in Algorithm \ref{alg:boosting} (Line 14). With $df$'s increase from 2 to $L$ (corresponding to $\lambda$'s decrease from $\infty$ to $0$), $b_{i}(y_{i};\gamma_{i,k})$ changes from a simple line fit to the ordinary least squares fit. For single weak learner, we only need to determine its hyper-parameter $df$ beforehand, simply choosing the small value (slightly greater than 2, and our method's default value is 3) for $df$ to restrict the model complexity (slighter more complex than the simple line fit). In fact, when $df$ is fixed with the default value, the number of boosting iterations $M$ is the only hyper-parameter in the proposed method. Fortunately, it\cite{Bartlett1998} has been shown that even when the training error is close to zero (log-likelihood is close to the upper-bound in our algorithm), boosting provably tends to increase the performance on the testing data with the increase of $M$. We later provide robust experiment to validate this surprising phenomenon.

\begin{algorithm}[!t]  
	\caption{Estimating the density distribution of $y_{i}$ by boosting NPMLE}  
	\label{alg:boosting}  
	\begin{algorithmic}[1]
		\STATE{Initialization}  
		\STATE {$b_{i}(y_{i};\gamma_{i,0})\gets -\frac{1}{2}y_{i}^{2}-\frac{1}{2}\ln\,2\pi$}
		\STATE {$f_{i}^{0}(y_{i})\gets b_{i}(y_{i};\gamma_{i,0})$}
		\STATE {$f_{i}^{0^{'}}(y_{i})\gets b_{i}^{'}(y_{i};\gamma_{i,0})$}
		\STATE {$f_{i}^{0^{''}}(y_{i})\gets b_{i}^{''}(y_{i};\gamma_{i,0})$}
		\STATE {compute the $L$ grid values $\{y_{i}^{*l}\}_{l=1}^{L}$ and $\Delta_{i}$ from $\{y_{i}^{j}\}_{j=1}^{N}$. }
		\STATE {$\omega_{l}^{0}\gets \Delta_{i} \frac{1}{\sqrt{2\pi}}e^{-\frac{y_{i}^{*l^{2}}}{2}},\,1\le l \le L$}
		
		\FOR {$k = 1$ to $M$}          
		\FOR {$l = 1$ to $L$}      
		\STATE {$\omega_{l}^{k} \gets\omega_{l}^{k-1}e^{{b_{i}(y_{i}^{*l};\gamma_{i,k-1})}}$}
		\STATE {$Y_{l}^{k} \gets\frac{q_{l}-\omega_{l}^{k}}{\omega_{l}^{k}}$}
		\ENDFOR
		\STATE{solve the reweighted least squares with respect to $\gamma_{i,k}$}
		\STATE {$\sum_{l=1}^{L}\frac{1}{2}\omega_{l}^{k}(b_{i}(y_{i}^{*l};\gamma_{i,k})-Y_{l}^{k})^{2} +\lambda J(b_{i}(y_{i};\gamma_{i,k}))$}
		\STATE{$f_{i}^{k}(y_{i})\gets f_{i}^{k-1}(y_{i})+b_{i}(y_{i};\gamma_{i,k})$}
		\STATE{$f_{i}^{k^{'}}(y_{i})\gets f_{i}^{k-1^{'}}(y_{i})+b_{i}^{'}(y_{i};\gamma_{i,k})$}
		\STATE{$f_{i}^{k^{''}}(y_{i})\gets f_{i}^{k-1^{''}}(y_{i})+b_{i}^{''}(y_{i};\gamma_{i,k})$}
		\ENDFOR
		\STATE{Output}
		\STATE{$f_{i}(y_{i})\gets f_{i}^{M}(y_{i})$}
		\STATE{$f_{i}^{'}(y_{i})\gets f_{i}^{M^{'}}(y_{i})$}
		\STATE{$f_{i}^{''}(y_{i})\gets f_{i}^{M^{''}}(y_{i})$}
	\end{algorithmic}  
\end{algorithm} 

\begin{figure}[h]
	\centering
	\includegraphics[width=0.5\linewidth]{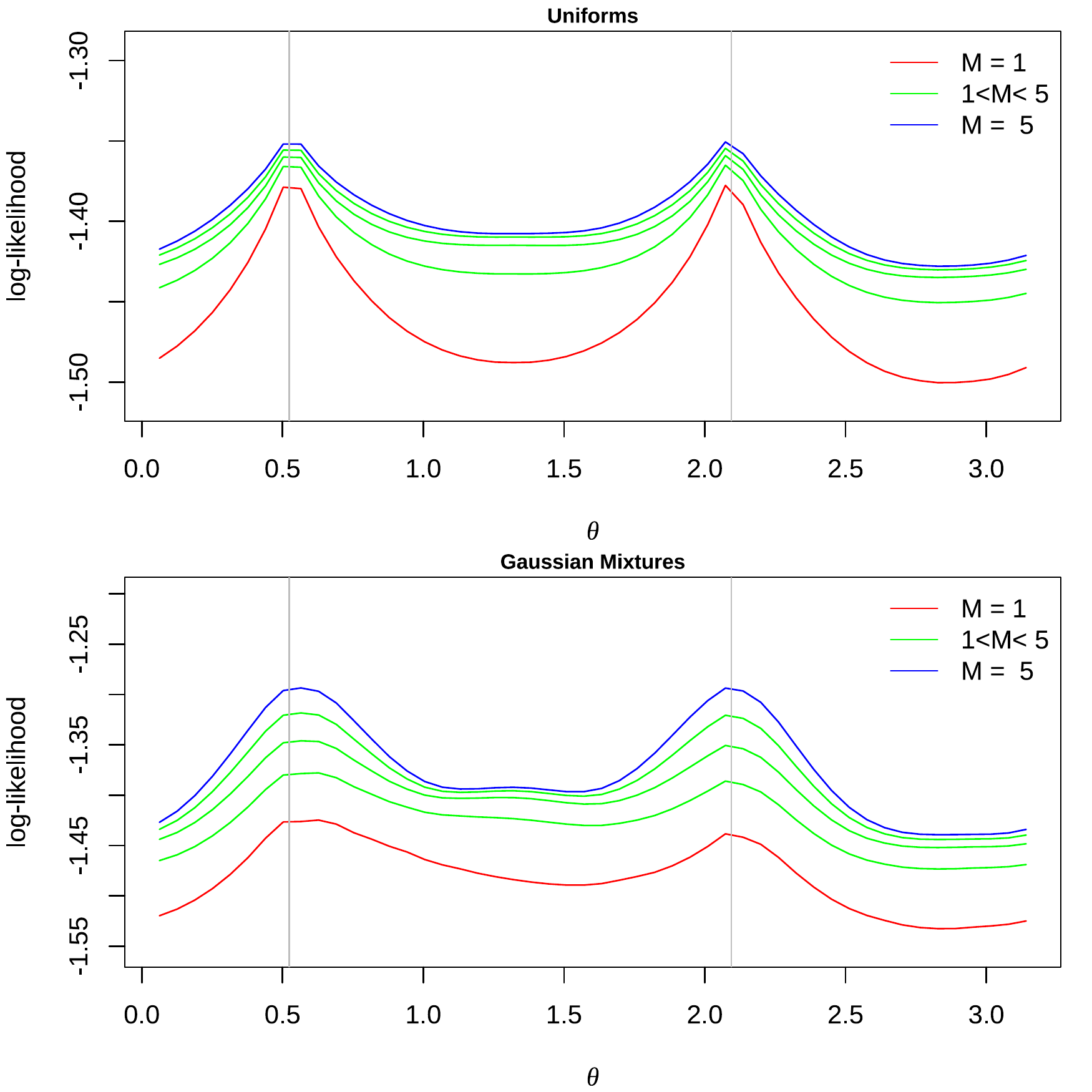}
	\caption{The log-likelihood estimations (smooth spline as weak learner, $df=3$) in two dimension using different boosting iterations $M$.
		The independent components are uniforms in the top figure and Gaussian mixtures in the bottom figure. In each figure, the ordinate is the log-likelihood estimations and the abscissa represents the rotate angles $\theta$ of $\mathbf{W}$. The vertical lines are shown to indicate true rotate angles.}
	\label{loglikelihood2dim}
\end{figure}

Figure \ref{loglikelihood2dim} shows the estimated log-likelihood, and the true rotate angles (concerning the mixing matrix $\mathbf{A}$) of mixtures $\mathbf{x}$ are plotted for comparison. Smooth spline surprisingly performs well when $M = 1$, and it successfully finds the ground-truth in two cases as the increase of boosting iterations $M$.

\subsection{Fixed-point method for estimating the separation matrix}
\label{sec2_2}
Given the fixed $f_{i}$, partition function becomes constant. Since there exits a whitening preprocessing stage for the observation x and the unmixing matrix $\mathbf{W}$ is orthonormal, fixed-point method developed in $\mathit{FastICA}$\cite{fastica1999} can be used in our algorithm. Then, the log-likelihood is maximized by an approximative Newton method with quadratic convergence (see Algorithm \ref{alg:fixed-point}). 

\begin{algorithm}[!t]  
	\caption{Estimating the unmixing matrix $\mathbf{W}$ by fixed-point method}  
	\label{alg:fixed-point}  
	\begin{algorithmic}[1]
		\FOR {$i = 1$ to $m$}             
		\STATE {$\mathbf{w}_{i} \gets \mathrm{E}\{\mathbf{x}f^{'}_{i}(\mathbf{w}_{i}^{T}\mathbf{x})\}-\mathrm{E}\{f^{''}_{i}(\mathbf{w}_{i}^{T}\mathbf{x})\}\mathbf{w}_{i}$}
		\ENDFOR
		\STATE{$\mathbf{W}\gets(\mathbf{w}_{1},\cdots,\mathbf{w}_{m})^{T}$}
		\STATE{symmetric decorrelation}
		\STATE{$\mathbf{W}\gets(\mathbf{W}\mathbf{W}^{T})^{-\frac{1}{2}}\mathbf{W}$}
	\end{algorithmic}  
\end{algorithm}


\subsection{Discussions of BoostingICA}
\label{sec2_3}

Since there is a trade-off between $df$ and $M$ in boosting, simple hyper-parameters tuning strategies are available:
\begin{itemize}
	\item decreasing the value of boosting iterations $M$ might lead to the reduce of elapsed time, at the cost of weakening the density estimation for unknown sources.
	\item  slightly increasing the weak learner's model complexity $df$ might benefit the density estimation without sacrificing time efficiency.
	\item we can level up the separation performance by slightly increasing $df$, and reduce the elapsed time via cutting down $M$. 
\end{itemize}

Owing that $s_{i}$ is zero mean and unit variance, the most uncertain (or circumspect) initialization (Line 2) in Algorithm \ref{alg:boosting} is the logarithm of the standard Gaussian,  which is different from the zero initialization in the original boosting NPMLE\cite{ypl2021}.

\section{Experiments and Results}
\label{sec3}
\subsection{Implementation details}
\label{sec3_1}


 Our experiments were conducted on the R development platform with Intel(R) Core(TM) i5-8250U CPU @ 1.60GHz, and we intend to illustrate the proposed method's comparable performance with other popular or recent ICA algorithms.

ICA algorithms used in our experiments are shown in Table \ref{table:methods}. $FixNA2$ is the recent blind source separation algorithms based on nonlinear auto-correlation, and $EFICA$ \cite{Kol2006} is a statistically efficient version of the $FastICA$. All methods share the same maximum iterations $maxit=20$, and $EFICA$'s elapsed time (with superscript *) was measured on the Matlab platform while other methods were R implementations.

The separation performance is measured by the value of Amari metrics $d(\mathbf{W},\mathbf{W}_{0})$ \cite{Amari1999},
\begin{equation}
\label{Amarimetric}
\begin{aligned}
&d(\mathbf{W},\mathbf{W}_{0})\\
&=\frac{1}{2m}\sum_{i=1}^{m}\left(\frac{\sum_{j=1}^{m}|r_{ij}|}{\max_{j}|r_{ij}|}-1\right)
+\frac{1}{2m}\sum_{j=1}^{m}\left(\frac{\sum_{i=1}^{m}|r_{ij}|}{\max_{i}|r_{ij}|}-1\right)
\end{aligned}
\end{equation}
where $r_{ij}=(\mathbf{W}\mathbf{W}_{0}^{-1})_{ij}$, $\mathbf{W}_{0}$ is the known truth. $d(\mathbf{W},\mathbf{W}_{0})$ is equal to zero if and only $\mathbf{W}$ and $\mathbf{W}_{0}$ are equivalence.
Besides Amari metrics, we also used the signal-to-interference ratio SIR ($SIR$ function in $\emph{ica}$ package\cite{icapackage}) as the criterion, where larger SIR indicates better performance in ICA.

\begin{table}[]
	\caption{ICA methods used in the experiments.}
	\label{table:methods}
	\centering
	\begin{tabular}{llll} 
		\toprule
		Methods&Symbols  &Parameters\\ 
		\midrule
		FastICA(G0)\cite{fastica1999,proDenICApackage}&F-G0  &$G_{0}(u)=\frac{1}{4}u^{4}$ \\
		FastICA(G1)\cite{proDenICApackage}&F-G1  &$G_{1}(u)=\ln \cosh(u)$  \\
		ProDenICA\cite{HaTi2002,proDenICApackage}&PICA  & default \\
		FixNA2\cite{mati2017,tsBSS}&FNA2  & default  \\
		BoostingICA(smooth spline)&B-SP  &$L=500$,\quad $df=3$\quad$M=5$    \\
		EFICA\cite{Kol2006,EFICA} &EICA& default\\
		\bottomrule
	\end{tabular}
\end{table}

\subsection{Audio separation task}
Two kinds of sources were used in the audio separation experiment: 6 speech recordings were from male (MJ60\_07, MA03\_01, MJ57\_03) and female (FC14\_04,FC18\_06, FD19\_06) speakers in TSP dataset \cite{kabal2002tsp}, and 18 different random sources were from the past researches\cite{Bach2003,HaTi2002}.
These 24 sources ($N=15200$) were mixed by an invertible
matrix to produce the high dimensional observed mixtures $\mathbf{x}$, and such procedure was replicated 50 times.

\begin{table}[]
	\caption{The Amari metrics (multiplied by 100), SIR and CPU elapsed time for Audio separation task (50 replications).}
	\label{table:audio}
	\begin{tabular}{lllllll}
		\toprule
		Mean& F-G0  & F-G1  & PICA & B-SP & FNA2& EICA \\
		\midrule
		Amari metrics& 30.51 & 30.64 & \textbf{12.41} & 13.67 &414.86 & 29.53\\
		SIR & 21.98 &23.64& \textbf{30.14}& 29.30& 7.02& 27.16\\
		Elapsed time (s) &1.73 & 1.35 & \textbf{17.54} & 20.51 & 381.15 & $0.71^{*}$\\
		\bottomrule
		\\
		\toprule
		Standard deviation& F-G0  & F-G1  & PICA & B-SP & FNA2& EICA\\
		\midrule
		Amari metrics& 0.00 &4.14 &0.64& 1.01& 22.20& 4.56\\
		SIR &  0.00 & 0.48 & 0.35 &0.41& 0.22 &0.71 \\
		Elapsed time (s) &0.11& 0.10& 0.38& 0.69& 23.69& $0.01^{*}$\\
		\bottomrule
	\end{tabular}
\end{table}
As can be seen from the Table \ref{table:audio}, B-SP, PICA, EICA acquired the best three separation performances, and
$FNA2$ was actually not converged. Although B-SP performed slightly better than PICA in this experiment, we will later illustrate the simple way to improve our methods' separation performance and time efficiency.
\subsection{Natural scene images separation task}
\label{sec3_3}
We designed an images separation experiment in this subsection, where the three gray-scale images were chosen from the \emph{ICS} \cite{nordhausen2008tools} package. These images depict a forest road, cat and sheep, and they have been used in many ICA researches \cite{nordhausen2008tools}. We vectorized them to arrive into a $130^{2}\times 3$ data matrix and we fixed the mixing matrix as $\mathbf{A}=\left[0.8,0.3,-0.3;0.2,-0.8,0.7;0.3,0.2,0.3\right]$ (column-wise).


The results are recorded in Table \ref{ICSexp}, where PICA performed better than B-SP ($df=3, M=5$) both in Amari/SIR and elapsed time. To improve the performance of B-SP, we have to level up its separation performance and cut down its elapsed time at the same time. 
Following the instructions in Subsection \ref{sec2_3}, we successfully found the appropriate tuning parameters $(df=8,M=3)$ for B-SP in few attempts, and we show them at the bottom of Table \ref{ICSexp}. 
\begin{table}[h]
	\caption{The Amari metrics (multiplied by 100), SIR and CPU elapsed time for image separation task.}
	\label{ICSexp}
	\centering
	\begin{small}
		\begin{tabular}{llll}
			\toprule
			methods& Amari metrics 	&SIR&Elapsed time (s) \\
			\midrule
			F-G0 & 38.61 & 10.69 &   0.24\\
			F-G1 & 54.88 & 9.04  & 0.17\\
			PICA &19.04 &16.39  & 1.95\\
			B-SP ($df=3,M=5$) &24.45 &14.55 &2.62\\
			FNA2 &32.17 &11.84 &0.44\\
			EICA  &37.87 &10.11  &$0.24^{*}$\\
			\midrule
			B-SP ($df=5,M=3$) &19.90 &15.91  &1.65\\
			B-SP ($df=8,M=3$) &\textbf{18.73} &\textbf{16.56} &  \textbf{1.66}\\
			\bottomrule
		\end{tabular}
	\end{small}
\end{table}

\begin{figure}[!h]
	\centering
	\centerline{\includegraphics[width=0.6\linewidth]{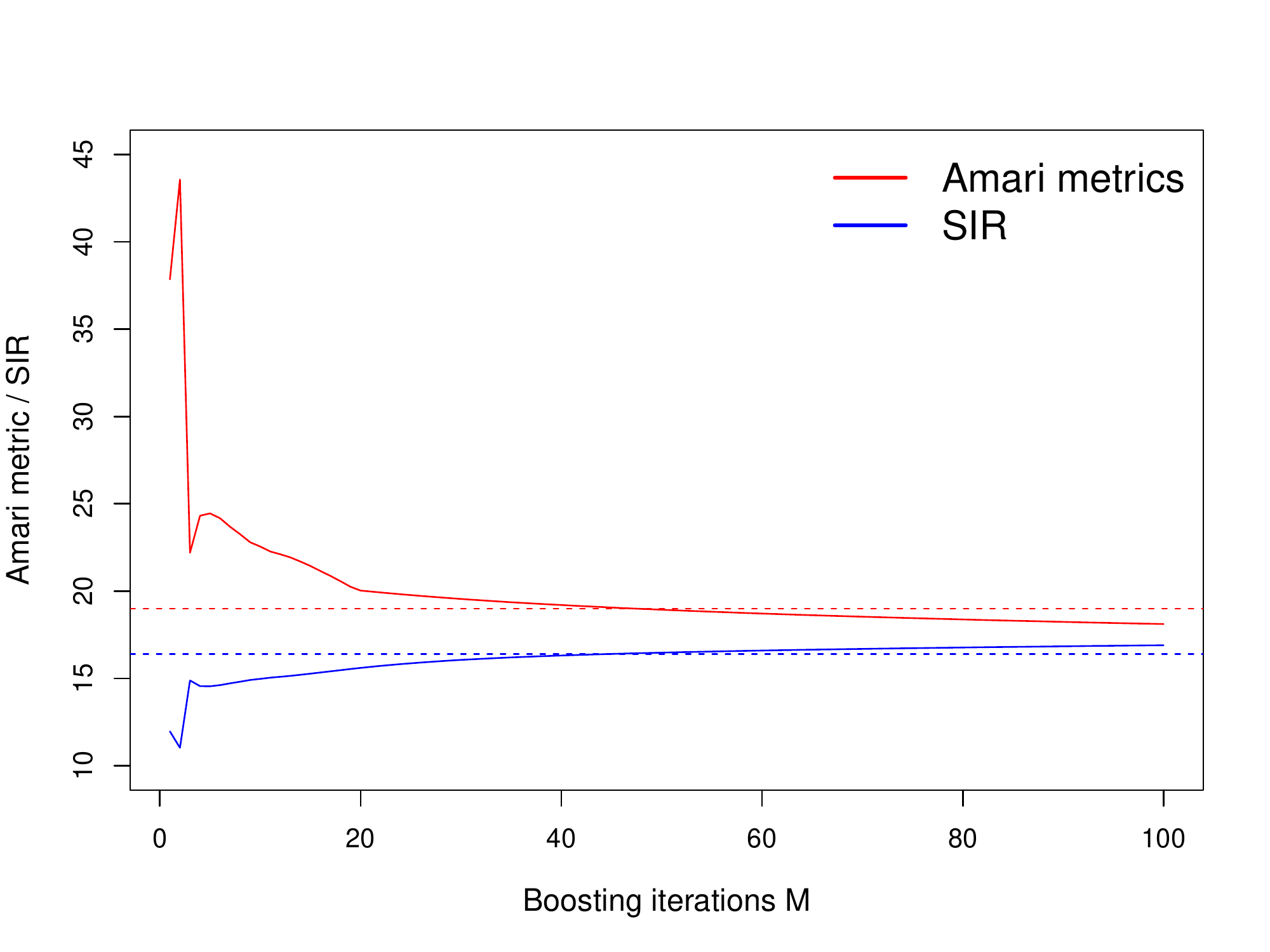}}
	\caption{ Separation performances of B-SP on \emph{ICS} images mixtures. The ordinate is the Amari metrics/SIR, and the abscissa represents boosting iterations $M$. The performance of PICA is shown in the dot line for comparison, and B-SP overwhelmed PICA when $M\ge48$.}
	\label{ICSexp_robust}
\end{figure}

Once the running time is not our key interest, it might be an appropriate way to improve the separation performance via simply increasing $M$. The corresponding robust experiments on \emph{ICS} data are shown in Figure \ref{ICSexp_robust}, and B-SP outperformed PICA when $M\ge48$.

\section{Conclusion}
\label{sec4}
In this paper, we introduce boosting to the nonparametric independent component analysis to alleviate the density mismatching between unknown sources and their estimations. The proposed algorithm is based on our earlier research \cite{ypl2021}, and it is a joint likelihood maximization between boosting NPMLE and fixed-point method. The proposed method $BoostingICA$ \footnote{code available: yunpli2sp.github.io}is concise and efficient, a list of experiments have illustrated its competitive performance with other popular or recent ICA algorithms.



%



\ifCLASSOPTIONcaptionsoff
  \newpage
\fi

\bibliographystyle{IEEEtran}
\bibliography{ref2}

\end{document}